\documentclass{article}

\usepackage[nonatbib,final]{nips_2016}

\usepackage[utf8]{inputenc} 
\usepackage[T1]{fontenc}    
\usepackage{hyperref}       
\usepackage{url}            
\usepackage{booktabs}       
\usepackage{amsfonts}       
\usepackage{nicefrac}       
\usepackage{microtype}      
\usepackage{xcolor}

\usepackage{graphicx}
\usepackage{caption}
\usepackage{subcaption}
\usepackage{algorithm}
\usepackage{algorithmic}
\usepackage{amsmath}

\title{DeepSpark: A Spark-Based Distributed Deep Learning Framework for
Commodity Clusters}

\author{
  Hanjoo Kim, Jaehong Park, Jaehee Jang, and Sungroh Yoon \\
  Electrical Engineering and Computer Science \\
  Seoul National University \\
  Seoul 08826, Korea \\
  \texttt{sryoon@snu.ac.kr} \\
}

\begin{document}

\maketitle

\begin{abstract}
The increasing complexity of deep neural networks (DNNs) has made it challenging
to exploit existing large-scale data processing pipelines for handling massive
data and parameters involved in DNN training. Distributed computing platforms and
GPGPU-based acceleration provide a mainstream solution to this computational
challenge. In this paper, we propose DeepSpark, a distributed and parallel deep
learning framework that exploits Apache Spark on commodity clusters. To support
parallel operations, DeepSpark automatically distributes workloads and
parameters to Caffe/Tensorflow-running nodes using Spark, and iteratively
aggregates training results by a novel lock-free asynchronous variant of the popular
elastic averaging stochastic gradient descent based update scheme, effectively
complementing the synchronized processing capabilities of Spark. DeepSpark is an
on-going project, and the current release is available at \url{http://deepspark.snu.ac.kr}.
\end{abstract}

\section{Introduction}
Deep neural networks (DNNs) continue to push the boundaries of their application
territories. For instance, convolutional neural networks (CNNs) have become the
\emph{de facto} standard method for image/object recognition in computer vision~\cite{he2015deep}. Other types of DNNs have also
shown outstanding performance in various machine learning problems including
speech recognition~\cite{chorowski2015attention} and image
classification~\cite{he2015deep,simonyan2014very}.

DNNs deliver a sophisticated modeling capability underpinned by multiple hidden
layers (\emph{e.g.}, a recent CNN model called ResNet consists of over 150
layers~\cite{he2015deep}), which effectively provide intermediate representations of the original input data. Leveraged by
this condition, DNNs can better handle complications in machine learning
applications, compared to previous techniques. Although having multiple hidden
layers allows DNNs to have powerful non-linear modeling capability, training such DNNs generally requires a large volume
of data and a huge amount of computational resources. This leads to long
training time ranging from several hours to days even, with general purpose
graphics processing unit (GPGPU) based
acceleration~\cite{he2015deep,simonyan2014very,szegedy2015going,krizhevsky2012imagenet}.

Various approaches have been proposed to improve the efficiency of deep
learning training. Highly optimized GPGPU implementations have significantly
shortened the time spent on training DNNs, often showing $10$--$100\times$
speed-up~\cite{chetlur2014cudnn,krizhevsky2014one}.
However, accelerating DNN training on a single machine has limitations because
of the limited resources such as GPU memory or the host machine's main
memory~\cite{krizhevsky2012imagenet}. Scaling out methods in distributed
environments have been
suggested~\cite{li2014scaling,dean2012large,ho2013more,xing2015petuum,ooi2015singa}
to overcome such issues. These approaches
exploit data parallelism and/or model parallelism and can potentially provide scalability.

On the other hand, a seamless integration of DNN training with existing data
processing pipelines is also an important practical point to avoid unnecessary
transfer and duplication. Many real-world datasets used for DNN training (such
as raw images or speech signals) need to be converted into a format required by
deep learning platforms and often require preprocessing to improve robustness
~\cite{krizhevsky2012imagenet, geiger2014investigating}. Such datasets are
typically huge in scale, thus the preprocessing procedure demands a considerable
amount of time and resources and requires carefully designed software to
process.
Inspired by the needs, SparkNet ~\cite{moritz2015sparknet} and
CaffeOnSpark\footnote{\url{https://github.com/yahoo/CaffeOnSpark}} combined deep
learning algorithms with existing data analytic pipelines on Apache
Spark~\cite{zaharia2010spark}. However, these implementions seldom achieve
speed-up for the cluster built on commodity hardware infrastructure, since Spark
lacks useful techniques and optimizations for DNN training.

We propose DeepSpark, a new deep learning framework on Spark, to accelerate DNN
training, and address the issues encountered in large-scale data handling.
Specifically, our contributions include the following:
\begin{enumerate}
    \item \textit{Seamless integration of scalable data management capability with deep learning:} We implemented our deep
    learning framework interface and the parameter exchanger on Apache Spark,
    which provides a straightforward but effective data parallelism layer.

    \item \textit{Overcoming high communication overhead using asynchrony:} We
    implemented an asynchronous stochastic gradient descent (SGD) for better DNN
    training in Spark. We also implemented an adaptive variant of the elastic
    averaging SGD (EASGD), which gave rise to faster parameter updates and
    improved the overall convergence rate, even on the low bandwidth network. Additionally,
    we described the speed-up analysis for our parallelization scheme.
	\item \textit{Flexibility:} DeepSpark supports Caffe and TensorFlow, two
	popular deep learning frameworks for accelerating deep network training. To the best of the authors' knowledge, this is the first attempt to integrate TensorFlow with Apache Spark.
	\item \textit{Availability:} The proposed DeepSpark library is freely available at \url{
	http://deepspark.snu.ac.kr}.
\end{enumerate}

\begin{figure}[t]
\centering
\begin{minipage}{.55\textwidth}
  \centering
  \includegraphics[width=.95\linewidth]{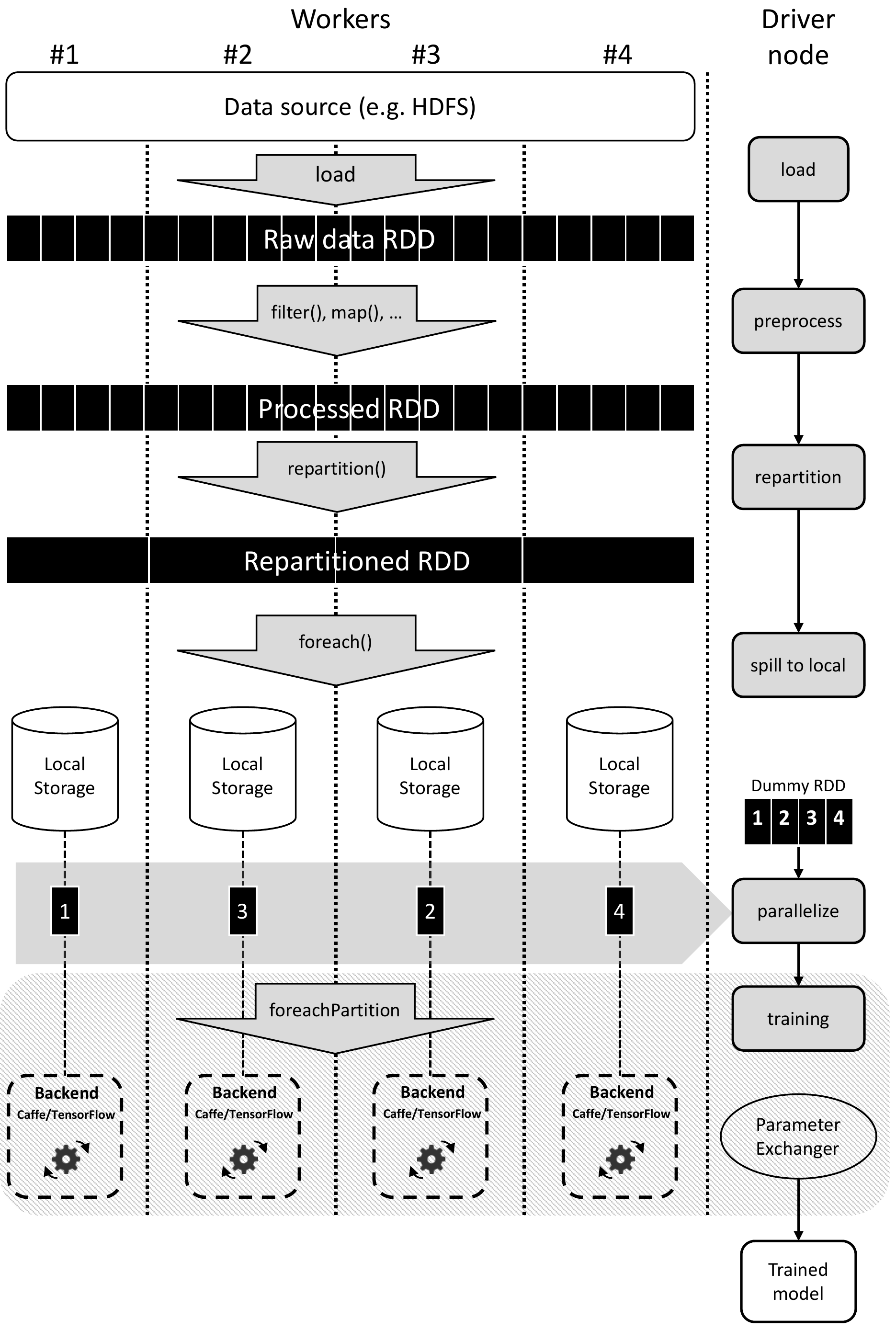}
  \captionof{figure}{DeepSpark training workflow}
  \label{fig:workflow}
\end{minipage} \qquad
\begin{minipage}{.37\textwidth}
  \centering
  \includegraphics[width=.95\linewidth]{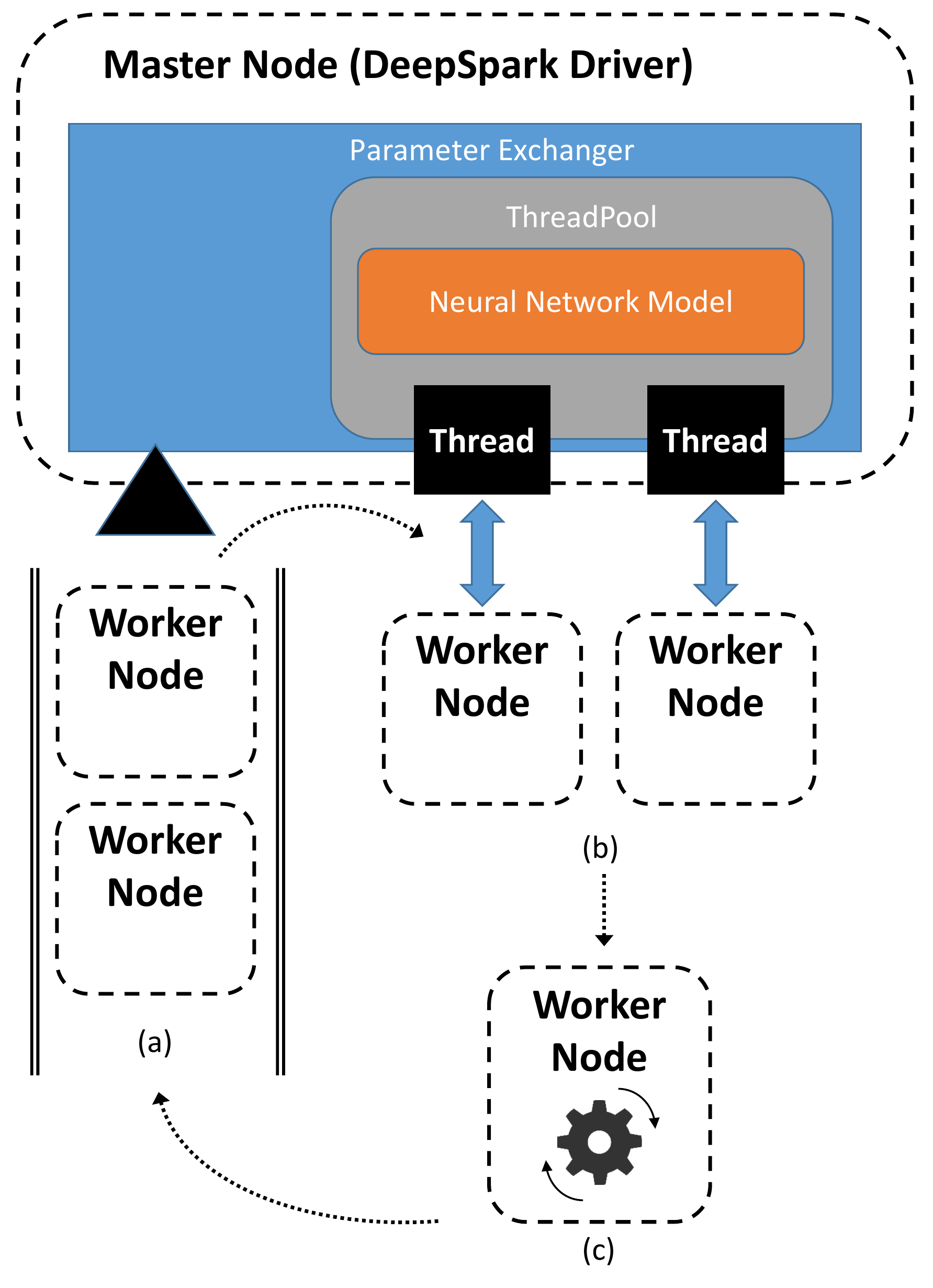}
  \captionof{figure}{Learning process with parameter exchanger. (a) Worker
nodes that want to exchange the parameters are waiting in a queue until
an available thread appears. (b) Exchanger threads take care of the worker request. In this example,
there are two exchanger threads in the pool. (c) After exchange, a
worker node performs its SGD process during the communication period $\tau$.}
  \label{fig:param}
\end{minipage}
\end{figure}

\section{Distributed Deep Learning}
Training DNN is composed of two steps: feed-forward and
backpropagation~\cite{hecht1989theory}. Feed-forward produces output using
previous activations and hidden parameters. The loss is then computed at the output classifier and is backpropagated to the
previous layer through the entire network. Parameter optimization (\emph{e.g.},
stochastic gradient descent) is executed during the backpropagation step in order to better
fit the target function. Although the complex neural network model
successfully approximates input data distribution, it inherently leads to a large amount of parameters to learn. This
situation demands a huge amount of time and computational resources, which are
two of the major concerns in deep learning research.

\subsection{Parallelizing SGD}
A na\"{i}ve parallelization SGD can be implemented by splitting batch calculations over
multiple worker nodes ~\cite{teo2007scalable, zinkevich2010parallelized}.
The global parameters are initialized and broadcasted from master to
each worker, and the workers will then derive gradients from their local data.
Considering that the throughput of each node can differ, two strategies can be
utilized to parallelize gradient update: synchronized or non-synchronized.

Synchronous SGD waits for every worker node to finish its computation and reach the barrier.
Once all worker nodes have completed their tasks, the master node collects all
the gradients, averages them, and applies them to the center parameter. Worker
nodes then pull the updated parameters from the master.
Although synchronous SGD is the most straightforward form of parallelizing SGD,
its performance is highly degraded by the slowest worker. It also suffers from
network overhead while aggregating and broadcasting parameters through
the network.

Asynchronous SGD has been suggested to resolve the inefficiency caused by the synchronous barrier
locking. In the lock-free asynchronous SGD, each worker node independently communicates with the central parameter server
without waiting for other nodes to finish. This strategy seems to be at risk for
stale gradients, however, asynchronous SGD has been theoretically and
empirically investigated to converge faster than the SGD on a single
machine~\cite{dean2012large,lian2015asynchronous, zhang2015deep}.

\subsubsection{Parameter Server}
The notion of A parameter server is a framework that aims for large-scale
machine learning training and inference~\cite{li2014scaling,ho2013more}. Its
master-slave architecture distributes data and tasks over
workers, and server nodes manage the global parameters. In previous works,
a parameter server has been successfully used in training various machine
learning algorithms, such as logistic regression and latent dirichlet allocation
~\cite{blei2003latent} on petabytes of real data.

\subsection{Reducing Communication Overhead}
The distributed training of DNNs consists of two steps:
exploration and exploitation~\cite{zhang2015deep}. The former is to explore the
parameter space to identify the optimal parameters by gradient descent,
and the latter is to update center parameters using local workers'
training results and proceed to the next step. Given that the parameter
exchanging causes network overhead, a performance tradeoff exists between
workers' exploration and a master's exploitation.

Some approaches proposed reducing the communication dominance instead of
accepting the penalty of the discrepancy. SparkNet presented the iteration
hyperparameter $\tau$, which is the number of processed minibatches before the next
communication~\cite{moritz2015sparknet}. The distributed training system can
benefit from the large value of $\tau$ under the high-cost communication
scenario by reducing communication overhead. However, large $\tau$ may end up
requiring more iterations to attain convergence, which slows down the learning
process~\cite{moritz2015sparknet,zhang2015deep}.

Zhang et al. (2015) suggested the EASGD strategy to maximize the benefit of
exploring~\cite{zhang2015deep} by regulating the discrepancy between a master
and workers. This method is different from downpour SGD~\cite{dean2012large}, where gradients of local
workers are shipped to the master, and updated center parameters are sent
back to workers at every update. When updating parameters, worker nodes compute
the elastic difference between them and apply the difference on both master and worker parameters.
To compute the elastic force, moving rate $\alpha \in (0,1)$ is involved. At
every communication, each worker and the master node update their parameters as
follows:
\begin{equation}
\begin{gathered}
	w_{worker} = w_{worker} - \alpha (w_{worker} - w_{master}) \\
	w_{master} = w_{master} + \alpha (w_{worker} - w_{master})
\end{gathered}
\label{eq:update}
\end{equation}
EASGD shows a faster convergence of the training even with large value of
$\tau$, with which downpour SGD shows a slow convergence rate or even cannot
converge~\cite{zhang2015deep}.

\section{DeepSpark Implementation}
\subsection{Motivations}
Apache Spark is an attractive platform for data-processing pipelines such as
database query processing. However, the frequent demand for communication
operates to the disadvantage of the synchronous SGD process on the commodity
environment. Furthermore, Spark RDD provides limited asynchronous operations
between the master and the workers.

To address the disadvantages of Spark, we implemented a new asynchronous SGD
solver with a custom parameter exchanger on the Spark environment.
Additionally, we designed the heuristic modification for EASGD
algorithm~\cite{zhang2015deep} by considering adaptive parameter updates
period to alleviate communication load.

DeepSpark consists of three main parts: namely, Apache Spark,
a parameter exchanger for asynchronous SGD, and the GPU-supported computing
engine. Apache Spark manages workers and available resources assigned by a resource manager. Figure~\ref{fig:workflow} depicts how the Spark workflow progresses for asynchronous SGD, and we provide more detailed descriptions in Section 3.3.
Subsequently, we explain the parameter exchanger and asynchronous SGD process
exploiting Spark in Sections 3.4--6 and Figure~\ref{fig:param}. We clarify how to
integrate Caffe~\cite{jia2014caffe} and
TensorFlow~\cite{tensorflow2015-whitepaper} with Spark using the Java Native
Access (JNA)\footnote{\url{https://github.com/java-native-access/jna}} interface and a heuristic method for the adaptive communication period in Section 3.6. The overall procedure of DeepSpark is summarized in Algorithm~\ref{alg:proposed}.

\begin{algorithm}[t]
   \caption{Pseudo-procedure of DeepSpark with adaptive communication period
   $\tau$ for $p$ workers and a master node.}
   \label{alg:proposed}
\begin{algorithmic}[1]
   \STATE {\bfseries Input:} communication period $\tau$, iterations $i_{max}$,
   learning rate $\eta$, data partition $D$ on HDFS, loss threshold
   $\text{L}_{\text{cut}}$, and moving average rate $\alpha$

   \STATE
   \STATE $D_{k} = \text{spillToLocal}(\text{Partition}_{D})$ // Initialize for
   the a worker node $k$

   \STATE
   \STATE $x_{k} = x_{\text{master}}$ // A variant of asynchronous EASGD update
   parameters

   \STATE
   \STATE $\text{L} = 0$
   \FOR{$i = 0$ to $i_{max}$}
   		\STATE $x_{k}^{(i+1)} = x_{k}^{(i)} - \eta \nabla f(x_{k}^{(i)};D_{k})$
   		\STATE $\text{L} = \text{L} + f(x_{k}^{(i)};D_{k})$
   		\STATE
   		\STATE // Adaptively update
   		\IF{$\text{L} > \text{L}_{\text{cut}}$}
   			\STATE $\text{EASGDUpdate}(x_{k}^{(i)},x_{\text{master}},\alpha)$
   			\STATE $\text{L} = 0$
   		\ENDIF
   \ENDFOR

\end{algorithmic}
\end{algorithm}

\subsection{Distributed Setup for Spark}
In this section, we explain DeepSpark's distributed workflow from the
data preparation to asynchronous SGD, which corresponds to line 3 in
Algorithm~\ref{alg:proposed}, and from load to spilling phase in
Figure~\ref{fig:workflow}. Given that DeepSpark is running on top of the Spark
framework, it needs to load and transform raw data in the form of Spark
RDD~\cite{zaharia2012resilient}.

The first step in DeepSpark training is to create RDD for training and
inference. We defined a container class, which stores label information and
corresponding data for each data sample. The data-specific loader then creates
the RDD of this data container class, which is followed by the preprocessing
phase. In the preprocessing phase, data containers would be connected to the
preprocessing pipeline such as filtering, mapping, transforming, or shuffling. The processed RDD
repartitioning is then performed to match the number of partitions
to the number of worker executors.

Caffe and TensorFlow, the actual backend computing engine, however, cannot
directly access RDD. In DeepSpark, the entire dataset is distributed across all
workers, and each worker can cache or convert its own parts of the dataset into the
LMDB\footnote{http://lmdb.readthedocs.org/en/release/} file format if the data
are relatively larger than the memory size. For a relatively small dataset, the
RDD \texttt{foreachPartition} action is executed, where every data partition is
loaded in the local worker's memory as a form of the Java List. These data then
become available to neural network model of the backend using a memory data
layer, or a tensor object for the Caffe and TensorFlow, respectively. In this
case, we should set the dimension of data, such as batch size, the number of
channels, the image width, and height.

The other approach to feed the data into the backend is spilling the dataset on
a worker node's storage. For a large dataset that is difficult to hold in the
physical memory, the RDD \texttt{foreach} operation is performed, and each data
partition is converted to the LMDB file format and stored in the temporary local
repository of the node it belongs to. Once the LMDB files are created, Caffe
automatically computes data dimension and finally completes the neural network
model parameters. We used LMDB JNI\footnote{https://github.com/chirino/lmdbjni}
to manipulate LMDB on Spark.

\subsection{Asynchronous EASGD Operation}
Inherently, Spark does not support step-wise asynchronous operations for
asynchronous SGD updates. We adopt the method that exploits Spark RDD
operations to overcome the limitation of Spark. The dummy RDD represented in
Figure~\ref{fig:workflow} can mimic the asynchrony.

Once the LMDB local repository for each worker has been prepared, the dummy RDD
is created and distributed across every worker. These dummy data have an
important role in launching the distributed action (\emph{i.e.}, parallel model
training is performed). Although the explicit dependency between spilling and training
steps is nonexistent at the code level, each worker node would be guided to
launch the training process with a spilled dataset by the dummy RDD. This
exploits the property of Spark that the Spark scheduler reuses the pre-existing worker
node session. The size of the dummy RDD is explicitly set to the number of
workers for full parallel operation, and the \texttt{foreachParition} action is
executed on this dummy RDD. Inside the \texttt{foreachPartition} process, each
worker can use the local data repository that has been created in the previous
job and starts the training step.

During the training process, the Spark driver program serves as a central
parameter exchanger, which performs an asynchronous EASGD update. At the initial
step, the driver node broadcasts its network address and neural network setup
files to all workers. Workers then create their own models and start training
using broadcasted data.

\subsection{Parameter Exchanger}
The parameter exchanger is the DeepSpark implementation of parameter server
concepts, which is essential for asynchronous update. In DeepSpark, the
application driver node serves as the parameter exchanger to enable worker nodes
to update their parameters asynchronously. Figure~\ref{fig:param} shows the
outline of the learning cycle with the parameter exchanger. The driver node
starts the parameter exchanger as a separate thread before worker nodes begin training the
model.

When multiple-parameter exchange requests from worker nodes exist, a thread pool
is implemented to handle the requests at the same time. For each connection
request, the thread pool allocates the pre-created threads that process the
parameter-exchange requests. The size of the thread pool is fixed in the program
, and we set this up to eight threads because of limited memory and network
bandwidth. If the number of requests exceeds the size, the unallocated requests
wait in a queue until the preceding requests are completed as shown in
Figure~\ref{fig:param}(a).

Exchange threads asynchronously access and update the neural net model in the
parameter exchanger based on the EASGD algorithm. In
Figure~\ref{fig:param}(b), given that it is a lock-free system, parameters can
be overwritten by simultaneous updates. Nevertheless, training results
are accumulated successfully, as proven in \cite{recht2011hogwild}. After the
parameter exchange action, each worker returns to the SGD phase to explore the parameter space
asynchronously as shown in Figure~\ref{fig:param}(c).

\subsection{Backend Engine and Adaptive Communication Period}
Each worker node in DeepSpark can use either the Caffe library 
and TensorFlow for the GPU-accelerated backend engine. However, Spark application is written in Java,
Scala, and Python, which cannot use the native backend engines directly in the
source code level. We implemented our code with JNA so that 
Spark executors can reference the native library of Caffe and TensorFlow.
The original \texttt{SGDSolver} class of Caffe does not provide an interface for that. Thus, we derived a custom solver class from the \texttt{SGDSolver}. We defined some operations of the derived solver
class to perform an atomic iteration action, acquire current trained parameters, and modify. This custom solver class
provides an interface to control the Caffe library for the DeepSpark application.
Therefore, current Caffe model prototype can be used in a distributed
environment without changing the Caffe network specifications numerous times. In case of
TensorFlow, we wrote an additional Java wrapper for TensorFlow C++ native
library. We can run a Tensorflow session on the wrapper with the operational
graph which was obtained from model description written in Python.

Because we observed that the testing loss decreased faster at the early learning
stage, we guessed that the early learning stage communication affected more the
model training than the communication of later iteration.
Thus, we designed the heuristics to increase the communication period as the
model trained. At the initial setup stage, we set the threshold to communicate.
During the training process, each executor cumulates its own training loss. At
the time the cumulated loss arrives at the threshold, the executor performs
parameter exchanging and then resets the cumulated loss.
In a general situation, the communication period will increase as the
training loss drops. The grown period aids in relaxing communication overhead.
Line 13 in Algorithm~\ref{alg:proposed} corresponds to determining whether
to update or not.

\begin{figure}[t!]
\centering
\includegraphics[scale=.75]{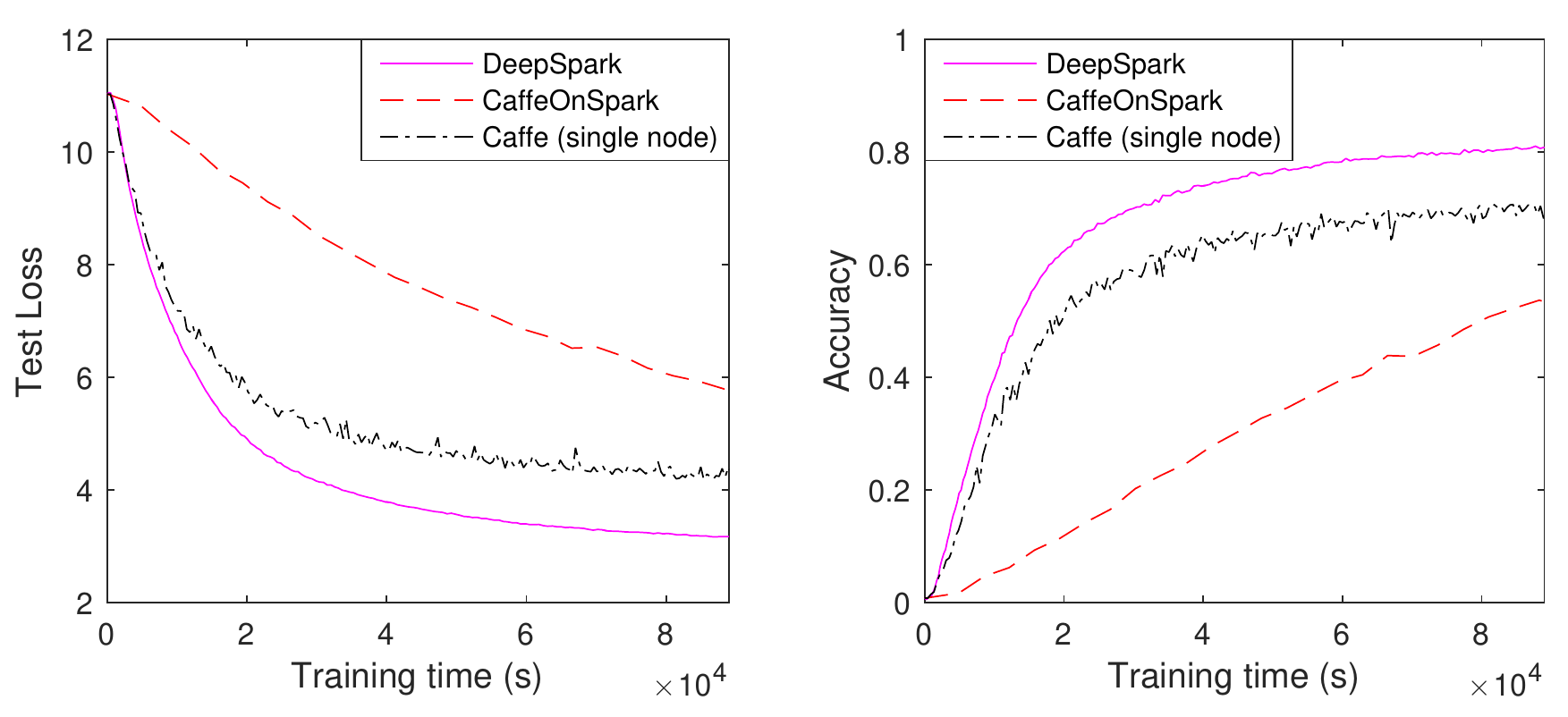}
\caption{ImageNet training results on DeepSpark, CaffeOnSpark, and Caffe
(original). Testing loss versus training time (left) and testing top-5 (right)
accuracy versus training time. DeepSpark and CaffeOnSpark were tested on 16
executors.}
\label{fig:imagenet_result}
\end{figure}

\section{Experiments}
\subsection{Experimental Setup}
We prepared a single-machine environment and a distributed cluster environment.
The distributed cluster was composed of 25 machines which are identical, and one
of them was used for the experiments on a single machine.
Each machine had an Intel Core i7-4790 processor with 16GB of main memory, and
an NVIDIA GTX970 GPU with 4GB memory and they communicated with each other via
Gigabit Ethernet (1 GigE) connections. We examined not only DeepSpark on the
cluster, but also Caffe~\cite{jia2014caffe} (single machine) and CaffeOnSpark (cluster), as 
sequential and parallel opponents in a day, respectively.

To manage each node, we set up an additional machine as well for a Hadoop
manager. This server machine did not participate in computing but only ran a
YARN resource manager daemon that performs as a manager of cluster nodes and an
HDFS namenode daemon.

The ILSVRC 2012 ImageNet dataset consists of 1,281,167 color
images with 1,000 classes of different image concepts~\cite{ILSVRC15}. As
pre-processing, we unified the size of images ($256 \times 256$), converted them
into a readable format for Apache Spark, and saved them on HDFS. To train the
ImageNet dataset, we used a Caffe model that is a replication~\footnote{\url{https://github.com/BVLC/caffe/tree/master/models/bvlc_googlenet}}
of Google's GoogLeNet~\cite{szegedy2015going}.
GoogLeNet is a 22-layer deep network made with the Hebbian principle and
multi-scale processing intuition, with nine of its inception modules for
classification. We sampled 5,000 images, which are not used in the training
stage for testing the trained models.

\subsection{Experimental Results}
We observed the training tendencies of three different methods on ImageNet: a
single Caffe machine, DeepSpark, and CaffeOnSpark. All the experiments were
performed in the same configuration, which includes learning rate $\eta=0.05$,
weight decay 0.002, and moving rate $\alpha=0.1$. The communication period
parameters $\tau$ for DeepSpark were selected from $\{100,200,500\}$ and the
adaptive setting. Figure~\ref{fig:imagenet_result} shows the accuracy and testing loss
versus training time. DeepSpark converged faster and more accurately than Caffe
on a single node. For the maximum achievements of each framework within the
training time, DeepSpark showed speed-up by 6.0 at 0.6 accuracy against
CaffeOnSpark, and 2.5 at 0.7 accuracy against single-node Caffe.
DeepSpark also attained a higher test accuracy (by 9\%) than Caffe on a
single node after converging. CaffeOnSpark did not reach the
converged state within the given training time.

\begin{figure}[t]
	\begin{minipage}[b]{.32\textwidth}
	  \centering
	  \includegraphics[width=\textwidth]{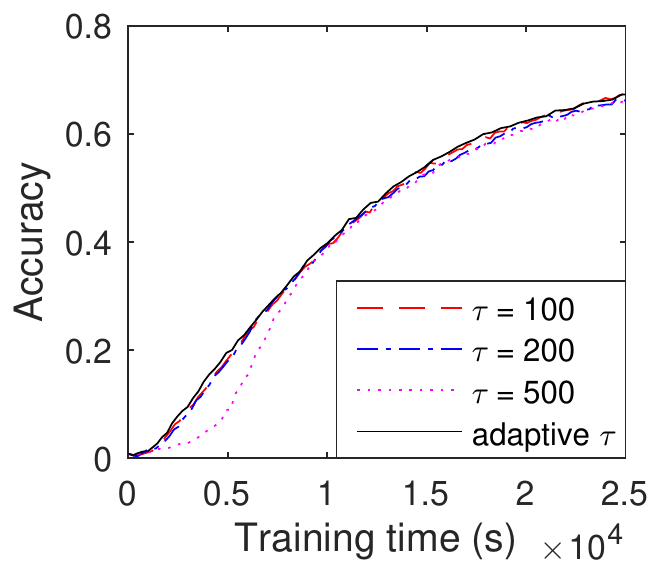} 
	  \captionof{figure}{Training results on different $\tau$.}
	  \label{fig:tau}
	\end{minipage} \hfill
 	\begin{minipage}[b]{.35\textwidth}
 	  \centering
 	  \includegraphics[width=\textwidth]{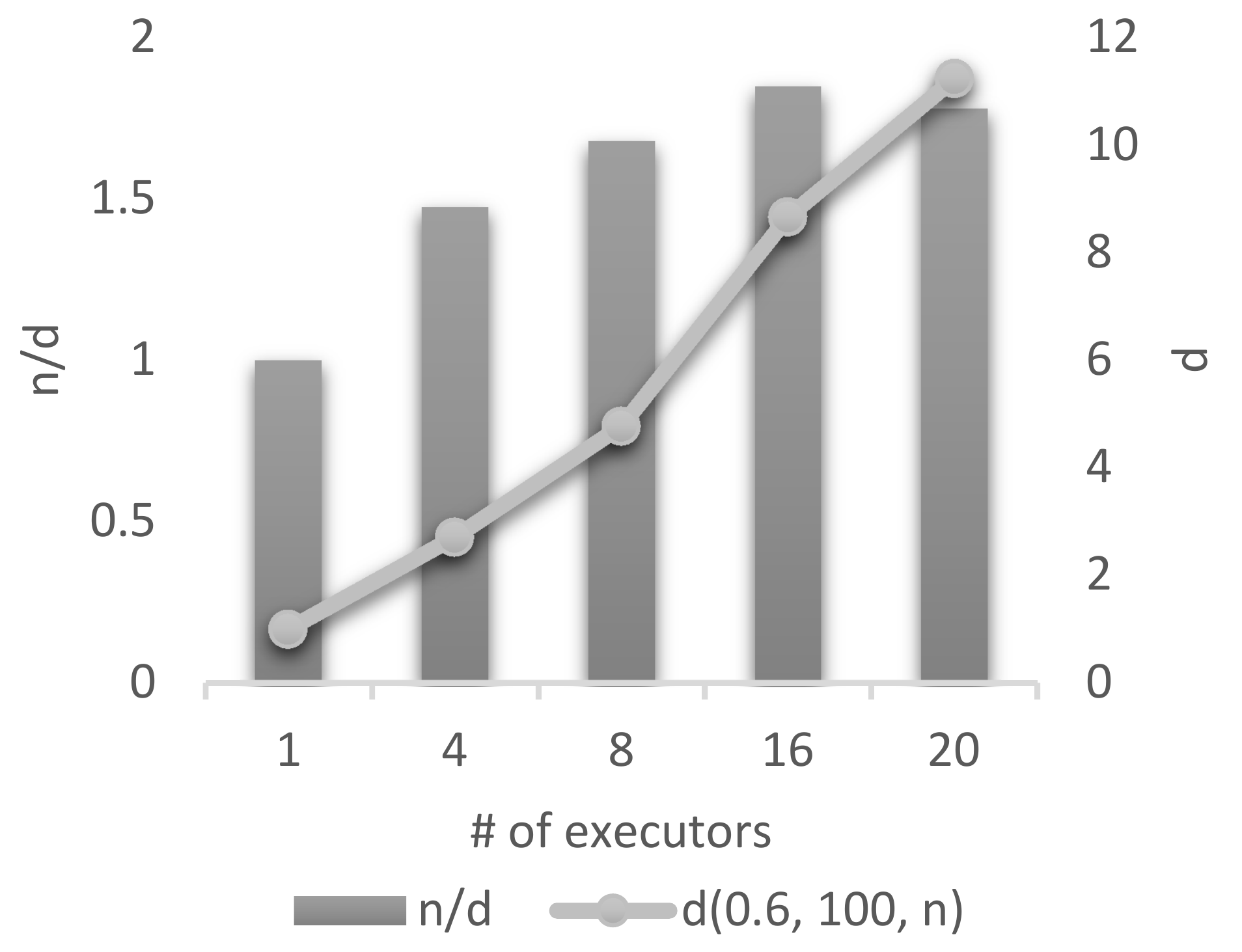}
 	  \captionof{figure}{The number of executors $n$ versus the $d(0.6,100,n)$ and
 	  the ratio $n/d(a,\tau,n)$.}
 	  \label{fig:d_dn}
 	\end{minipage} \hfill
	\begin{minipage}[b]{.27\textwidth}
	  \centering
	  \begin{tabular}{cc} \hline
	    $\tau$    & $d(0.6,\tau,16)$ \\ \hline
        100 & 8.66 \\
		200 & 9.21 \\
		500 & 9.64 \\ \hline
	  \end{tabular}
	  \captionof{table}{$d(0.6,\tau,16)$ on the different $\tau$. As the $\tau$
	  grows, the discrepancy penalty increases in a small step.}
	  \label{tbl:d_tau}
	\end{minipage}
\end{figure}

We measured the turnaround time per training 1,000 iterations, for
comparing the communication overhead of DeepSpark and CaffeOnSpark. DeepSpark
spent approximately 250 seconds to processs the iterations on all 16 executors
with $\tau=200$, and CaffeOnSpark consumed about one hour. The adaptive-period
EASGD showed slightly better performance than the fixed-period
EASGD setting at the early training stage, as shown Figure~\ref{fig:tau}.

\section{Discussion}
\subsection{Comparison between DeepSpark and CaffeOnSpark}
DeepSpark achieved a high speed-up compared with the sequential Caffe and the
distributed CaffeOnSpark by alleviating communication overhead. Communication
overhead played a crucial role in the slowdown. We attempted to relax the communication overhead $S$ by
a delayed asynchronous update and succeeded in accelerating DNN training. The
time DeepSpark spent to aggregate and broadcast was less than 1/10 of the time
CaffeOnSpark spent. Taking GoogLeNet training on DeepSpark (16 nodes), for
example, the time spent in parameter exchange is about $10 \times$ of the time
spent in minibatch training.

In terms of the training epoch, the penalty of discrepancy caused by
the delayed asynchronous update may double the number of required iterations to
achieve the accuracy. Each executor needs 32,000 runs to achieve 0.6 accuracy on
CaffeOnSpark, while 84,000 iterations were required to achieve the same level of
accuracy for each DeepSpark executor. Consequently, since the benefit of
reducing communication overhead surpass the penalty of discrepancy, DeepSpark gained the
speed-up against CaffeOnSpark.

\subsection{Speed-up Analysis}
We describe the computation and communication time, $T_{comp}$ and $T_{comm}$,
from DeepSpark's parallelization scheme as the following,
\begin{equation}
\label{eq:time}
	T_{comp} = N_{a}(b)C(b)\frac{d(a,\tau,n)}{n}, \quad T_{comm} =
	\frac{nd(a,\tau,n)N_{a}(b)}{\tau}S
\end{equation}
where $N_{a}(b)$ is the required number of iterations to achieve the
target accuracy level $a$, $b$ is the size of the minibatch, $C(b)$ is the time
for a batch computation, $S$ is a communication overhead, $n$ represents the number of
nodes, $\tau$ is a communication period, and $d(a,\tau,n)$ implies a discrepancy
penalty by asynchrony. We supposed that the discrepancy
penalty $d_{\tau,n}$ has a positive correlation with $\tau$ and $n$ by the
result. From Equation~(\ref{eq:time}), we can draw the
speed-up function as follows,
\begin{equation}
\label{eq:speedup}
\begin{gathered}
	\text{Speed-up} = \frac{N_{a}(b)C(b)}{T_{comp} + T_{comm}}
	= \frac{n \tau}{\tau d(a,\tau,n) + \frac{d(a,\tau,n)S}{C(b)}n^{2}}
\end{gathered}
\end{equation}

In case of the commodity cluster ($S >> C(b)$), the synchronous schemes
($\tau=1, d=1$) like CaffeOnSpark cannot make any speed-up even with the
infinite number of worker nodes, as shown in Figure~\ref{fig:imagenet_result}.
We can select large $\tau$ to achieve speed-up for that condition. For sufficiently
large $\tau$, Equation~(\ref{eq:speedup}) is approximated to
Equation~(\ref{eq:apr_speedup}) by neglecting the second term in the denominator, and
it is an theoretical speed-up for $a$,$\tau$,and $n$.
\begin{equation}
\label{eq:apr_speedup}
	\text{(\ref{eq:speedup})} \approx \frac{n}{d(a,\tau,n)}
\end{equation}

For the dataset that is too large to hold in memory, the distributed
environment was helpful in reducing the disk-operation load.
In the distributed system, the dataset is divided into $K$ shards,
and that shard of data may reduce disk I/O. In the results,
the disk I/O overhead was estimated approximately an hour per 32 batch size and
100,000 iterations in a single node. The overhead decreased by 42\% for 16
nodes.

Although we spilled the shard of data into local storage in our experiments,
The time delay caused by spilling did not affect the overall performance of
DeepSpark. On 16 executors, approximately 10 minutes were spent to spill
the ImageNet dataset, respectively. These delays accounted for just under
1\% of the entire running time, which was negligible.

\subsection{Limitations}
Intuitively, the discrepancy penalty $d(a,\tau,n)$ is expected to be dependent on
$a$, and has a positive correlation with $\tau$ and $n$, as shown in
Figure~\ref{fig:d_dn} and Table~\ref{tbl:d_tau}. The increament of
$d(a,\tau,n)$ was more influenced by the number of executors than by the 
communication period in our experimental setting. This will limit the
scalability, in terms of speed-up, as shown Figure~\ref{fig:d_dn}. Also, the broken lineage of RDD may weaken the fault tolerance of Spark by local spilling.
If an executor fails, the entire training process should be restarted in the current
implementation.

\section{Conclusion}
We have described our new deep learning framework, DeepSpark, which
provides seamless integration with existing large-scale data processing
pipelines as well as accelerated DNN training procedure. DeepSpark is an
example of a successful combination of diverse components including Apache
Spark, asynchronous parameter updates, and a GPGPU-based backend
computation engine.
Based on our experiments with popular benchmarks, DeepSpark demonstrated its
effectiveness by showing faster convergence than the alternative parallelization
schemes on the commodity cluster.

\bibliographystyle{unsrtnew}
\bibliography{document}

\begin{thebibliography}{10}

\bibitem{he2015deep}
K.~He, et~al.
\newblock Deep residual learning for image recognition.
\newblock {\em arXiv preprint arXiv:1512.03385}, 2015.

\bibitem{chorowski2015attention}
J.~K. Chorowski, et~al.
\newblock Attention-based models for speech recognition.
\newblock In {\em NIPS}, pages 577--585, 2015.

\bibitem{simonyan2014very}
K.~Simonyan et~al.
\newblock Very deep convolutional networks for large-scale image recognition.
\newblock {\em arXiv preprint arXiv:1409.1556}, 2014.

\bibitem{szegedy2015going}
C.~Szegedy, et~al.
\newblock Going deeper with convolutions.
\newblock In {\em CVPR}, pages 1--9, 2015.

\bibitem{krizhevsky2012imagenet}
A.~Krizhevsky, et~al.
\newblock Imagenet classification with deep convolutional neural networks.
\newblock In {\em NIPS}, pages 1097--1105, 2012.

\bibitem{chetlur2014cudnn}
S.~Chetlur, et~al.
\newblock cudnn: Efficient primitives for deep learning.
\newblock {\em arXiv preprint arXiv:1410.0759}, 2014.

\bibitem{krizhevsky2014one}
A.~Krizhevsky.
\newblock One weird trick for parallelizing convolutional neural networks.
\newblock {\em arXiv preprint arXiv:1404.5997}, 2014.

\bibitem{li2014scaling}
M.~Li, et~al.
\newblock Scaling distributed machine learning with the parameter server.
\newblock In {\em OSDI}, pages 583--598, 2014.

\bibitem{dean2012large}
J.~Dean, et~al.
\newblock Large scale distributed deep networks.
\newblock In {\em NIPS}, pages 1223--1231, 2012.

\bibitem{ho2013more}
Q.~Ho, et~al.
\newblock More effective distributed ml via a stale synchronous parallel
  parameter server.
\newblock In {\em NIPS}, pages 1223--1231, 2013.

\bibitem{xing2015petuum}
E.~P. Xing, et~al.
\newblock Petuum: A new platform for distributed machine learning on big data.
\newblock In {\em SIGKDD}, KDD '15, pages 1335--1344, New York, NY, USA, 2015.
  ACM.

\bibitem{ooi2015singa}
B.~C. Ooi, et~al.
\newblock Singa: A distributed deep learning platform.
\newblock In {\em Proceedings of the ACM International Conference on
  Multimedia}, pages 685--688. ACM, 2015.

\bibitem{geiger2014investigating}
J.~T. Geiger, et~al.
\newblock Investigating nmf speech enhancement for neural network based
  acoustic models.
\newblock In {\em INTERSPEECH}, pages 2405--2409, 2014.

\bibitem{moritz2015sparknet}
P.~Moritz, et~al.
\newblock Sparknet: Training deep networks in spark.
\newblock {\em arXiv preprint arXiv:1511.06051}, 2015.

\bibitem{zaharia2010spark}
M.~Zaharia, et~al.
\newblock Spark: cluster computing with working sets.
\newblock In {\em Proceedings of the 2nd USENIX Conference on Hot Topics in
  Cloud Computing}, volume~10, page~10, 2010.

\bibitem{hecht1989theory}
R.~Hecht-Nielsen.
\newblock Theory of the backpropagation neural network.
\newblock In {\em IJCNN}, pages 593--605. IEEE, 1989.

\bibitem{teo2007scalable}
C.~H. Teo, et~al.
\newblock A scalable modular convex solver for regularized risk minimization.
\newblock In {\em SIGKDD}, pages 727--736. ACM, 2007.

\bibitem{zinkevich2010parallelized}
M.~Zinkevich, et~al.
\newblock Parallelized stochastic gradient descent.
\newblock In {\em NIPS}, pages 2595--2603, 2010.

\bibitem{lian2015asynchronous}
X.~Lian, et~al.
\newblock Asynchronous parallel stochastic gradient for nonconvex optimization.
\newblock In {\em NIPS}, pages 2719--2727, 2015.

\bibitem{zhang2015deep}
S.~Zhang, et~al.
\newblock Deep learning with elastic averaging sgd.
\newblock In {\em NIPS}, pages 685--693, 2015.

\bibitem{blei2003latent}
D.~M. Blei, et~al.
\newblock Latent dirichlet allocation.
\newblock {\em the Journal of machine Learning research}, 3:993--1022, 2003.

\bibitem{jia2014caffe}
Y.~Jia, et~al.
\newblock Caffe: Convolutional architecture for fast feature embedding.
\newblock In {\em Proceedings of the ACM International Conference on
  Multimedia}, pages 675--678. ACM, 2014.

\bibitem{tensorflow2015-whitepaper}
M.~Abadi, et~al.
\newblock {TensorFlow}: Large-scale machine learning on heterogeneous systems,
  2015.
\newblock Software available from tensorflow.org.

\bibitem{zaharia2012resilient}
M.~Zaharia, et~al.
\newblock Resilient distributed datasets: A fault-tolerant abstraction for
  in-memory cluster computing.
\newblock In {\em Proceedings of the 9th USENIX Conference on Networked Systems
  Design and Implementation}, pages 2--2. USENIX Association, 2012.

\bibitem{recht2011hogwild}
B.~Recht, et~al.
\newblock Hogwild: A lock-free approach to parallelizing stochastic gradient
  descent.
\newblock In {\em NIPS}, pages 693--701, 2011.

\bibitem{ILSVRC15}
O.~Russakovsky, et~al.
\newblock {ImageNet Large Scale Visual Recognition Challenge}.
\newblock {\em IJCV}, 115(3):211--252, 2015.

\end{thebibliography}

\end{document}